\documentclass[11pt, oneside]{article}   	
\usepackage[margin=.8in]{geometry}               		
\usepackage{graphicx}				
\usepackage{amssymb}
\usepackage{color}
\usepackage{eqnarray, amsmath, amsthm}
\usepackage{appendix}
\usepackage{enumerate}
\usepackage{bbm}
\usepackage{mathabx}
\usepackage{hyperref}
\usepackage{authblk}
\usepackage[numbers]{natbib}

\newtheorem*{conjecture}{Conjecture}

\usepackage{tikz}
\usetikzlibrary{arrows.meta}

\title{Scaling up deep neural networks: a capacity allocation perspective}
\author{Jonathan Donier\footnote{jdonier@spotify.com}}
\affil{Spotify}

\begin{document}
\maketitle

\begin{abstract}
Following the recent work on capacity allocation, we formulate the conjecture that the shattering problem in deep neural networks can only be avoided if the capacity propagation through layers has a non-degenerate continuous limit when the number of layers tends to infinity. This allows us to study a number of commonly used architectures and determine which scaling relations should be enforced in practice as the number of layers grows large. In particular, we recover the conditions of Xavier initialization in the multi-channel case, and we find that weights and biases should be scaled down as the inverse square root of the number of layers for deep residual networks and as the inverse square root of the desired memory length for recurrent networks.
\end{abstract}

\section{Introduction} 

Capacity analysis has been introduced in \cite{donier2018capacity} as a way to analyze which dependencies a linear model is focussing its modelling capacity on, when trained on a given task. The concept was then extended in \cite{donier2018nonlinear} to neural networks with non-linear activations, where capacity propagation through layers was studied. When the layers are residual (or differential), and in one limiting case with extremely irregular activations (which was called the \emph{pseudo-random} limit), it has been shown that capacity propagation through layers follows a discrete Markov equation. This discrete equation can then be approximated by a continuous Kolmogorov forward equation in the deep limit, provided some specific scaling relation holds between the network depth and the scale of its residual connections -- more precisely, the residual weights must scale as the inverse square root of the number of layers. Following \cite{balduzzi2017shattered}, it was then hypothesized that the success of residual networks lies in their ability to propagate capacity through a large number of layers in a non-degenerate manner. It is interesting to note that the inverse square root scaling mentioned above is the only scaling relation that leads to a non-degenerate propagation PDE in that case: larger weights would lead to shattering, while smaller ones would lead to no spatial propagation at all. 

In this paper, we take this idea one step further and formulate the conjecture that enforcing the right scaling relations -- i.e. the ones that lead to a non-degenerate continuous limit for capacity propagation -- is key to avoiding the shattering problem: we call this the \emph{neural network scaling conjecture}. In the example above, this would mean that the inverse square root scaling must be enforced if one wants to use residual networks at their full power.

In the second part of this paper, we use the PDE capacity propagation framework to study a number of commonly used network architectures, and determine the scaling relations that are required for a non-degenerate capacity propagation to happen in each case.


\section{Capacity allocation: a reminder} 

\paragraph{Linear models} The aim of capacity allocation analysis is to determine how a model's parameters are being allocated to the input space, once it has reached an optimal state. The theory was first developed in the context of linear models with an $L^2$ loss in  \cite{donier2018capacity}, where the concept of capacity was first defined and a general formula was given. According to the formula, the total capacity of a model (which equals its number of effective parameters) can be broken down onto a partition of the input space -- for example, a spatial partition. More specifically, the capacity allocated to some subspace $\mathcal{S}$ of the input space can be expressed as:

\begin{equation}\label{eq:capacity_linear}
\kappa(\mathcal{S}) = \|K^T S \|_F^2,
\end{equation}

\noindent where $S$ is an orthonormal basis of $\mathcal{S}$ and $K$ is the model's capacity matrix, which can be derived from the autocovariance matrix of the inputs and the model state (cf.  \cite{donier2018capacity}).

\paragraph{Non-linear models} This result was then generalized to non-linear neural network layers in \cite{donier2018nonlinear}. Under some hypotheses (in particular, a very irregular form for the activation function), a formula was derived for capacity allocation though an arbitrary number of layers. More specifically, it was shown that residual convolutional layers act on the capacity as a spatial diffusion operator, such that capacity propagation through a large number of layers is described by a Fokker-Planck equation. In particular, this diffusive behaviour leads to an effective receptive field that grows as the square root of the number of layers -- which has been well-documented in the literature \cite{luo2016understanding}. But most importantly, the finding of a diffusive propagation through layers allows one to abstract away from the expression of the capacity given in Eq. (\ref{eq:capacity_linear}), as the conditions for a non-degenerate propagation to happen can be studied on their own. This is the aim of the present paper.

\section{The neural network scaling conjecture} 

Following these findings, it was suggested in \cite{donier2018nonlinear} that the capacity framework could be used to explain the shattering phenomenon, and therefore the success of residual networks (a.k.a. ResNets \cite{he2016deep}) in deep learning\footnote{For a gradients perspective on this question, see \cite{balduzzi2017shattered}.}. In the limit where the number of layers in the network tends to infinity, some particular scaling was required to keep some  weights of order 1 in the equation that transforms output capacity into input capacity, and under this scaling the capacity propagation equation obeys a Fokker-Planck equation in the deep limit. It is tempting from these observations to formulate what we call the \emph{neural network scaling conjecture}:

\begin{conjecture}
A deep neural network can only avoid the shattering problem 
if its capacity propagation equation has a non-degenerate continuous limit as the number of layers tends to infinity. 
\end{conjecture}

If true, this imposes scaling relations between various network hyper-parameters that must hold for very deep network to allocate capacity in a non-degenerate manner and thus train efficiently. One such scaling was obtained in \cite{donier2018nonlinear} between the number of layers and the scale of the weights in residual networks: if $L\gg 1$ is the total number of layers in the network, then the residual weights need to scale as $1 / \sqrt{L}$ in order to obtain a non-degenerate PDE for capacity propagation. Other examples are discussed below.

\section{Analysis of neural network architectures}\label{sec:architecture}

In this section, we consider a number of architectures that are commonly used in deep learning and derive the corresponding expressions for capacity propagation. In some cases, the neural network scaling conjecture suggests some particular scaling relations that should hold to achieve non-degenerate spatial propagation. Throughout this part, we will use the following notations:

\paragraph{General notations:}

\begin{figure}[!t]

\def\layersep{0.5cm}
\def\numnodes{5}
\def\middlenode{3}
\def\nodesep{0.74}

\def\ipos{2}
\def \offset{-1*\nodesep}
\def \offsetb{11.5*\nodesep}

\definecolor{darkgreen}{RGB}{0, 102, 0}

\newcommand\mydots{\vdots}

\hspace*{-0.5cm}
\begin{tikzpicture}[shorten >=1pt,->,draw=black!60, node distance=\layersep]
    \tikzstyle{every pin edge}=[<-,shorten <=1pt]
    \tikzstyle{neuron}=[circle,draw=black!80,minimum size=10pt,inner sep=0pt,line width=0.2mm]
    \tikzstyle{input neuron}=[neuron,fill=yellow!50];
    \tikzstyle{invisible neuron}=[neuron,minimum size=0pt];
    \tikzstyle{output neuron}=[neuron,fill=red!60];
    \tikzstyle{hidden neuron}=[neuron];
    \tikzstyle{annot} = [ text centered]

    \foreach \name / \y in {1,...,\numnodes}
        \node[input neuron] (I-\name) at (\offset-\nodesep*\y, 0) {};

    \foreach \name / \y in {1,...,\numnodes}
        \node[hidden neuron] (H1-\name) at (\offset-\nodesep*\y, \layersep) {};

    \foreach \name / \y in {1,...,\numnodes}
        \node[invisible neuron] (Z1-\name) at (\offset-\nodesep*\y, 2*\layersep) {};       
          
    \foreach \name / \y in {1,...,\numnodes}
        \node[invisible neuron] (Z3A-\name) at (\offset-\nodesep*\y, 5*\layersep) {};
        
    \foreach \name / \y in {1,...,\numnodes}
        \node[hidden neuron] (H3-\name) at (\offset-\nodesep*\y, 6*\layersep) {};

    \foreach \name / \y in {1,...,\numnodes}
        \node[invisible neuron] (Z3B-\name) at (\offset-\nodesep*\y, 7*\layersep) {};

    \foreach \name / \y in {1,...,\numnodes}
        \node[invisible neuron] (Z4A-\name) at (\offset-\nodesep*\y, 10*\layersep) {};
        
    \foreach \name / \y in {1,...,\numnodes}
        \node[hidden neuron] (H4-\name) at (\offset-\nodesep*\y, 11*\layersep) {};
               
    \node[output neuron,above of=H4-\middlenode, node distance=1cm] (O) {};


    \foreach \source in {1,...,\numnodes}
            \path[-] (I-\source) edge (H1-\source);
    \foreach \source[evaluate=\source as \dest using int(\source+1)] in {1,...,4}
            \path[-] (I-\source) edge (H1-\dest);
    \foreach \source[evaluate=\source as \dest using int(\source-1)] in {2,...,\numnodes}
            \path[-] (I-\source) edge (H1-\dest);

    \foreach \source in {1,...,\numnodes}
            \path[-] (H1-\source) edge (Z1-\source);
    \foreach \source[evaluate=\source as \dest using int(\source+1)] in {1,...,4}
            \path[-] (H1-\source) edge (Z1-\dest);
    \foreach \source[evaluate=\source as \dest using int(\source-1)] in {2,...,\numnodes}
            \path[-] (H1-\source) edge (Z1-\dest);

            \path[-] (Z1-\middlenode) -- node[auto=false]{\mydots} (Z3A-\middlenode);

    \foreach \source in {1,...,\numnodes}
            \path[-] (Z3A-\source) edge (H3-\source);
    \foreach \source[evaluate=\source as \dest using int(\source+1)] in {1,...,4}
            \path[-] (Z3A-\source) edge (H3-\dest);
    \foreach \source[evaluate=\source as \dest using int(\source-1)] in {2,...,\numnodes}
            \path[-] (Z3A-\source) edge (H3-\dest);
                        
    \foreach \source in {1,...,\numnodes}
            \path[-] (H3-\source) edge (Z3B-\source);
    \foreach \source[evaluate=\source as \dest using int(\source+1)] in {1,...,4}
            \path[-] (H3-\source) edge (Z3B-\dest);
    \foreach \source[evaluate=\source as \dest using int(\source-1)] in {2,...,\numnodes}
            \path[-] (H3-\source) edge (Z3B-\dest);
             
            \path[-] (Z3B-\middlenode) -- node[auto=false]{\mydots} (Z4A-\middlenode);            
    \foreach \source in {1,...,\numnodes}
            \path[-] (Z4A-\source) edge (H4-\source);
    \foreach \source[evaluate=\source as \dest using int(\source+1)] in {1,...,4}
            \path[-] (Z4A-\source) edge (H4-\dest);
    \foreach \source[evaluate=\source as \dest using int(\source-1)] in {2,...,\numnodes}
            \path[-] (Z4A-\source) edge (H4-\dest);
                                             
    \foreach \source in {1,...,\numnodes}
        \path[-] (H4-\source.north) edge (O.south);

    \node[annot] at (\offset - 6*\nodesep, 14.4 * \layersep) (legend) {Layer \#} ;

   \node[annot] (hl1)  at (\offset - 6*\nodesep, 0.5 * \layersep)  {$1$};
   \node[annot] (hll) at (\offset - 6*\nodesep, 6.5 * \layersep) {$\ell$};
   \node[annot] (hlL)  at (\offset - 6*\nodesep, 12 * \layersep) {$L$};

    \node[annot,below of=I-3, node distance=0.7cm] {Inputs};
    \node[annot,above of=O, node distance=0.7cm] {Output};
    
       \node[annot] (pi0) at (\offset+1*\nodesep, 11*\layersep) {$\pi(0, x)$}  ;
       \node[annot] (pit) at (\offset+1*\nodesep, 6*\layersep) {$\pi(t, x)$}  ;
       \node[annot] (pi1) at (\offset+1*\nodesep, 0*\layersep) {$\pi(1, x)$}  ;
    
      \%path[-] (pit) -- node[auto=false]{\mydots} (pi1);
    
    \node[annot] at (\offset + 0*\nodesep, 14.4 * \layersep) (legend) {Capacity} ;

   \path[line width=0.3mm, -{Latex[length=2mm,width=2mm]}] (\offset + 0*\nodesep, 13*\layersep) edge ( \offset +  0*\nodesep, -1*\layersep) ;
   
   \path[-, line width=0.3mm] (\offset + 0*\nodesep - 0.08, 0 * \layersep) edge (\offset +  0*\nodesep + 0.12, 0 * \layersep);
   \path[-, line width=0.3mm] (\offset +  0*\nodesep - 0.08, 6 * \layersep) edge (\offset +  0*\nodesep + 0.12, 6 * \layersep);
   \path[-, line width=0.3mm] (\offset +  0*\nodesep - 0.08, 11 * \layersep) edge (\offset +  0*\nodesep + 0.12, 11 * \layersep);

    \node[annot] at (\offset - \middlenode*\nodesep, -3.5* \layersep) (legend) {\begin{tabular}{c}(a) Network depth and capacity propagation\\ in  reverse layer time\end{tabular}} ;

      \path[-] (\offset + 4 *\nodesep, -1.5*\layersep) edge (\offset + 4 *\nodesep, 15 * \layersep);
		
%
%

    \foreach \name / \y in {1,...,\numnodes}
        \node[input neuron] (I-\name) at (\offsetb-\nodesep*\y, 0) {};

    \foreach \name / \y in {1,...,\numnodes}
        \node[hidden neuron] (H1-\name) at (\offsetb-\nodesep*\y, \layersep) {};

    \foreach \name / \y in {1,...,\numnodes}
        \node[invisible neuron] (Z1-\name) at (\offsetb-\nodesep*\y, 2*\layersep) {};       
          
    \foreach \name / \y in {1,...,\numnodes}
        \node[invisible neuron] (Z3A-\name) at (\offsetb-\nodesep*\y, 5*\layersep) {};
        
    \foreach \name / \y in {1,...,\numnodes}
        \node[hidden neuron] (H3-\name) at (\offsetb-\nodesep*\y, 6*\layersep) {};

    \foreach \name / \y in {1,...,\numnodes}
        \node[invisible neuron] (Z3B-\name) at (\offsetb-\nodesep*\y, 7*\layersep) {};

    \foreach \name / \y in {1,...,\numnodes}
        \node[invisible neuron] (Z4A-\name) at (\offsetb-\nodesep*\y, 10*\layersep) {};
        
    \foreach \name / \y in {1,...,\numnodes}
        \node[hidden neuron] (H4-\name) at (\offsetb-\nodesep*\y, 11*\layersep) {};
               
    \node[output neuron,above of=H4-\middlenode, node distance=1cm] (O) {};
    
    
  \node[hidden neuron, fill=blue!50] (H4fill-\ipos) at (\offsetb-\nodesep*\ipos, 11*\layersep) {};
  \node[hidden neuron, fill=darkgreen!50] (H3fill-\middlenode) at (\offsetb-\nodesep*\middlenode, 6*\layersep) {};


    \foreach \source in {1,...,\numnodes}
            \path[-] (I-\source) edge (H1-\source);
    \foreach \source[evaluate=\source as \dest using int(\source+1)] in {1,...,4}
            \path[-] (I-\source) edge (H1-\dest);
    \foreach \source[evaluate=\source as \dest using int(\source-1)] in {2,...,\numnodes}
            \path[-] (I-\source) edge (H1-\dest);

    \foreach \source in {1,...,\numnodes}
            \path[-] (H1-\source) edge (Z1-\source);
    \foreach \source[evaluate=\source as \dest using int(\source+1)] in {1,...,4}
            \path[-] (H1-\source) edge (Z1-\dest);
    \foreach \source[evaluate=\source as \dest using int(\source-1)] in {2,...,\numnodes}
            \path[-] (H1-\source) edge (Z1-\dest);

            \path[-] (Z1-\middlenode) -- node[auto=false]{\mydots} (Z3A-\middlenode);

    \foreach \source in {1,...,\numnodes}
            \path[-] (Z3A-\source) edge (H3-\source);
    \foreach \source[evaluate=\source as \dest using int(\source+1)] in {1,...,4}
            \path[-] (Z3A-\source) edge (H3-\dest);
    \foreach \source[evaluate=\source as \dest using int(\source-1)] in {2,...,\numnodes}
            \path[-] (Z3A-\source) edge (H3-\dest);
                        
    \foreach \source in {1,...,\numnodes}
            \path[-] (H3-\source) edge (Z3B-\source);
    \foreach \source[evaluate=\source as \dest using int(\source+1)] in {1,...,4}
            \path[-] (H3-\source) edge (Z3B-\dest);
    \foreach \source[evaluate=\source as \dest using int(\source-1)] in {2,...,\numnodes}
            \path[-] (H3-\source) edge (Z3B-\dest);
             
            \path[-] (Z3B-\middlenode) -- node[auto=false]{\mydots} (Z4A-\middlenode);            
    \foreach \source in {1,...,\numnodes}
            \path[-, draw=darkgreen!80, line width=0.4mm] (Z4A-\source) edge (H4-\source);
    \foreach \source[evaluate=\source as \dest using int(\source+1)] in {1,...,4}
            \path[-, draw=darkgreen!80, line width=0.4mm] (Z4A-\source) edge (H4-\dest);
    \foreach \source[evaluate=\source as \dest using int(\source-1)] in {2,...,\numnodes}
            \path[-, draw=darkgreen!80, line width=0.4mm] (Z4A-\source) edge (H4-\dest);
                                             
    \node[annot,below of=I-5, node distance=0.5cm] {$x_0$};
    \node[annot,below of=I-4, node distance=0.5cm] {$x_1$};
    \node[annot,below of=I-3, node distance=0.5cm] {$x_2$};
    \node[annot,below of=I-2, node distance=0.5cm] {$x_3$};
    \node[annot,below of=I-1, node distance=0.5cm] {$x_4$};
    
    \foreach \source in {1,...,\numnodes}
        \path[-, draw=darkgreen!80, line width=0.4mm] (H4-\source.north) edge (O.south);

     \node[annot] at (\offsetb - 6*\nodesep, 14.4 * \layersep) (legend) {Layer \#} ;

   \node[annot] (hl1)  at (\offsetb - 6*\nodesep, 0.5 * \layersep)  {$1$};
   \node[annot] (hll) at (\offsetb - 6*\nodesep, 6.5 * \layersep) {$\ell$};
   \node[annot] (hlL)  at (\offsetb - 6*\nodesep, 12 * \layersep) {$L$};
   
       \node[annot, color=blue] at (\offsetb+0.5*\nodesep, 11*\layersep) (legend) {$\pi(0, x_3)$}  ;
       \node[annot, color=blue] at (\offsetb+1.5*\nodesep, 10*\layersep) (legend) {$:= \kappa(L\to L, x_3)$}  ;
       \node[annot, color=darkgreen] at (\offsetb+ 0.5*\nodesep, 6*\layersep) (legend) {$\pi(t, x_2)$}  ;
       \node[annot, color=darkgreen] at (\offsetb+ 1.7*\nodesep, 5*\layersep) (legend) {$ := \kappa(L\to \ell(t), x_2)$}  ;
    
        \path[-, draw=blue!80, line width=0.4mm] (H4-\ipos.north) edge (O.south);
        \path[-, draw=darkgreen!80, line width=0.4mm] (H3-3) edge (Z3B-2);
        \path[-, draw=darkgreen!80, line width=0.4mm] (H3-3) edge (Z3B-3);
        \path[-, draw=darkgreen!80, line width=0.4mm] (H3-3) edge (Z3B-4);
        
            \node[annot,above of=O, node distance=0.7cm] {Output};
        
    \node[annot] at (\offsetb - \middlenode*\nodesep, -3* \layersep) (legend) {(b) Illustration of capacity propagation paths} ;
   
\end{tikzpicture}

\centering
\caption{\label{fig:schema} \textit{(left)} Network depth and propagation of the capacity $\pi(t, x) := \kappa(L\to \ell(t), x)$ from the last layer $L$ to lower layers, with $\ell(t) = L - t(L - 1)$. \textit{(right)} In blue: illustration of the capacity propagation path corresponding to $\kappa(L\to L, x_3)$ (in compact notation, $\pi(0, x_3)$). In green: capacity propagation paths corresponding to $\kappa(L\to \ell(t), x_2)$ (in compact notation, $\pi(t, x_2)$).}

\end{figure}
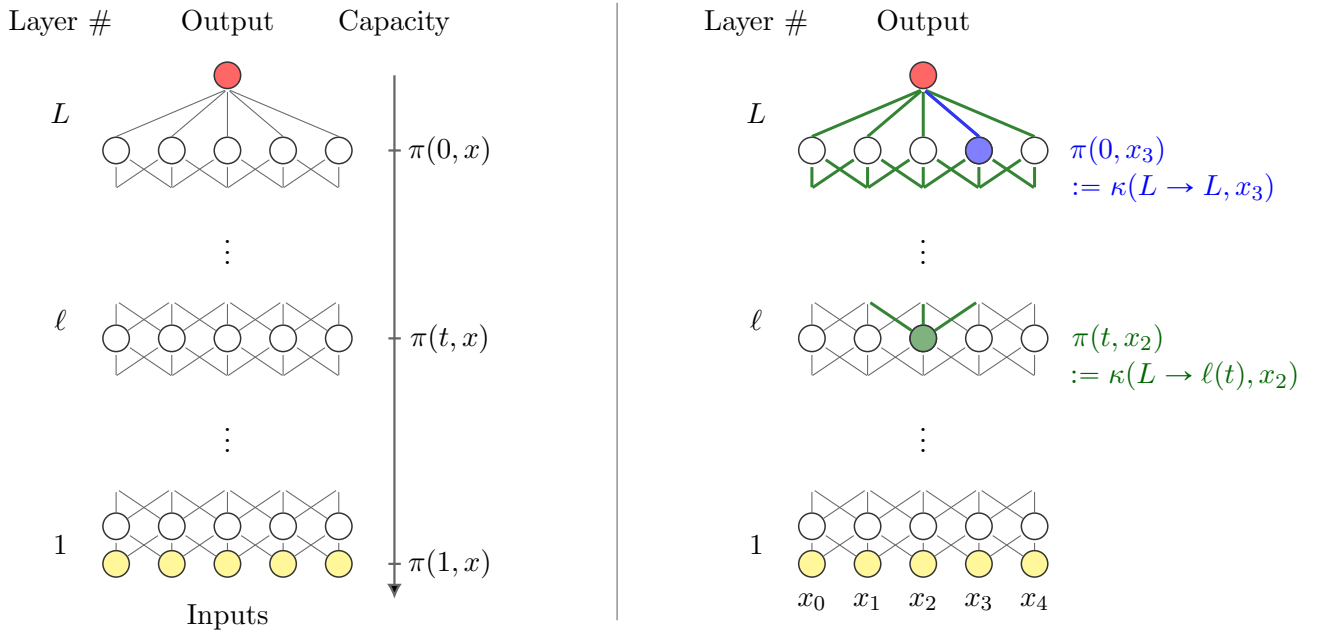

\begin{itemize}
\item $L$ is the number of layers in the network, where layers $1,\dots, L-1$ are hidden layers and layer $L$ is the (linear) output layer (see Fig. \ref{fig:schema}).
\item For $1\leq \ell \leq L$, the function performed by layer $\ell$ is noted $\phi_\ell$ (typically, an affine transformation followed by a non-linear activation).
\item The function performed by the ensemble of layers $1,\dots, L-1$ is noted $\phi$, such that $\phi(X)$ is the input of the last layer $L$ where $X$ is the input of the network.
\item $t:= (L - \ell) / (L - 1)$ is  the ``reverse layer time'', with $\ell$ being the index of a given layer in the network, such that $t=0$ for the output layer and $t=1$ for the input layer (see Fig. \ref{fig:schema}). Correspondingly, we note $\ell(t) = L - t(L - 1)$ the layer index that corresponds to a given reverse layer time $t\in[0, 1]$.
\end{itemize}

\paragraph{Capacity-related notations:}

\begin{itemize}
\item $\kappa(\ell'\to \ell, x)$ is the capacity allocated by the parameters of layer $\ell'$ to the inputs of layer $\ell \leq \ell'$. In particular, 
\begin{itemize}
\item[$\circ$] $\kappa(L\to L, x)$ (also noted $\kappa^L(x)$ or $\kappa^L$ for simplicity) is the capacity allocated by the linear output layer to its own inputs, as defined in \cite{donier2018capacity},
\item[$\circ$] $\kappa(L\to \ell, x)$ (also noted $\kappa^\ell(x)$ or $\kappa^\ell$ for simplicity) is the capacity allocated by the linear output layer to the inputs of hidden layer $\ell$, as derived in \cite{donier2018nonlinear} in the case of residual networks.
\end{itemize}
\item $\pi(t, x) := \kappa(L \to \ell(t), x)$ is a more compact notation to describe the capacity allocated by the last layer to layer $\ell(t)$. This reparametrization of depth in reverse layer time (between 0 at the output level and 1 at the input level) will be helpful when writing evolution PDEs below.
\end{itemize}

\subsection{Vanilla residual layers}

Before considering more complex architectures, let us start by recalling the capacity propagation PDE obtained in \cite{donier2018nonlinear}  for a residual network with pseudo-random activations. In this case, a single layer can be expressed as:

\begin{equation}
X_{\ell} = \phi_\ell(X_{\ell-1}) = f(W_\ell^TX_{\ell-1}),
\end{equation}

\noindent where $W_\ell$ is a square matrix. The capacity propagation from layer $\ell$ to layer $\ell-1$ is then described by the discrete Markov equation:

\begin{equation}
\kappa^{\ell-1} = \left(W_\ell \circ W_\ell\right) \kappa^\ell.
\end{equation}

\noindent If the weights $W_\ell$ are convolutional (hence local) and if the capacity propagation matrix can be written as $W_\ell \circ W_\ell = \mathbbm{1} + \epsilon \Delta_\ell$ with $\epsilon ~\propto ~1 / L$ (which is equivalent to scaling the off-diagonal weights of $W_\ell$ as $\epsilon^W ~\propto ~ 1 / \sqrt{L}$) then the above capacity propagation equation has the following continuous limit:

\begin{equation}\label{eq:capacity_PDE}
\begin{cases}
\displaystyle
\frac{\partial\pi}{\partial t}(t, x) = \mathcal{D}\frac{\partial^2 \pi}{\partial x ^2}(t, x),\\[6pt]
\pi(0, x) = \kappa^L(x),
\end{cases}
\end{equation}

\noindent with some diffusion coefficient $\mathcal{D}$ that depends on the statistics of the convolutional weights\footnote{Note that $\mathcal{D}$  can therefore depend on $t$, which we ignore here for simplicity.}, and where we have used the compact notation $\pi(t, x) := \kappa(L \to \ell(t), x)$ introduced in Section \ref{sec:architecture}. The solution to the above equation, and hence the propagated capacity $\kappa(L \to \ell, x)$, can expressed as an integral form:

\begin{equation}\label{eq:capacity_final_solution}
\kappa(L \to \ell(t), x) := \pi(t, x) =  \frac{1}{\sqrt{4\pi \mathcal{D}t}} \int_{-\infty}^\infty  e^{-\frac{(x-y)^2}{4\mathcal{D}t}} \kappa^L(y) \mathrm{d}y.
\end{equation}

\noindent As explained in  \cite{donier2018nonlinear}, this means in particular that the spatial propagation of capacity (and hence the size of the effective receptive field) scales as the square root of the number of layers, since the exponential factor in the above equation is of order unity for $x- y = O(\sqrt{\mathcal{D}t})$, with $t$ proportional to the number of layers traversed. Finally, note that the above equation was obtained under the following scaling:

\begin{equation}
\begin{cases}
\begin{aligned}
& \text{Residual capacity:} & \epsilon ~\propto ~1 / L \\
&\text{Residual weights:}& \epsilon^W := \sqrt{\epsilon} ~\propto~ 1 / \sqrt{L}\\
\end{aligned}
\end{cases}
\end{equation}

\subsection{Skip connections} \label{sec:vanilla}

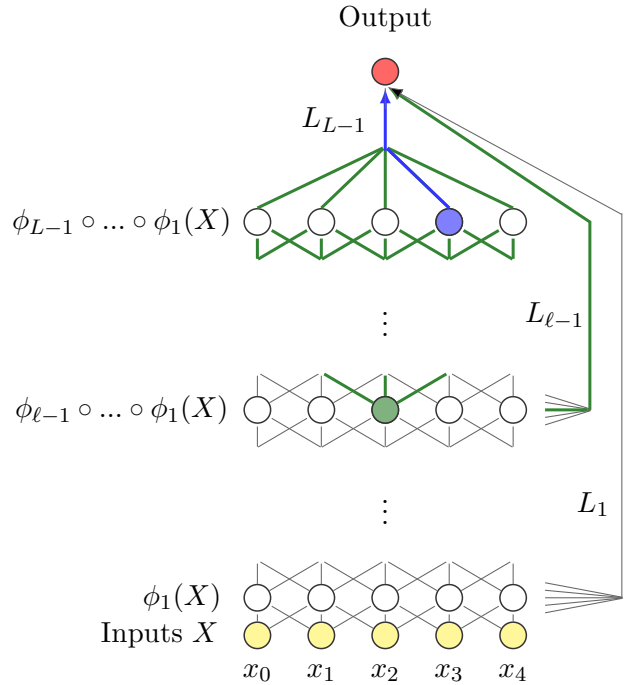
\begin{figure}[!b]

\def\layersep{0.5cm}
\def\numnodes{5}
\def\middlenode{3}
\def\nodesep{0.85}

\def\ipos{2}
\def \offsetb{0*\nodesep}

\definecolor{darkgreen}{RGB}{0, 102, 0}

\newcommand\mydots{\vdots}

\hspace*{4cm}
\begin{tikzpicture}[shorten >=1pt,->,draw=black!60, node distance=\layersep]
    \tikzstyle{every pin edge}=[<-,shorten <=1pt]
    \tikzstyle{neuron}=[circle,draw=black!80,minimum size=10pt,inner sep=0pt,line width=0.2mm]
    \tikzstyle{input neuron}=[neuron,fill=yellow!50];
    \tikzstyle{invisible neuron}=[neuron,minimum size=0pt];
    \tikzstyle{output neuron}=[neuron,fill=red!60];
    \tikzstyle{hidden neuron}=[neuron];
    \tikzstyle{annot} = [ text centered]

%
%

    \foreach \name / \y in {1,...,\numnodes}
        \node[input neuron] (I-\name) at (\offsetb-\nodesep*\y, 0) {};

    \foreach \name / \y in {1,...,\numnodes}
        \node[hidden neuron] (H1-\name) at (\offsetb-\nodesep*\y, \layersep) {};

    \foreach \name / \y in {1,...,\numnodes}
        \node[invisible neuron] (Z1-\name) at (\offsetb-\nodesep*\y, 2*\layersep) {};       
          
    \foreach \name / \y in {1,...,\numnodes}
        \node[invisible neuron] (Z3A-\name) at (\offsetb-\nodesep*\y, 5*\layersep) {};
        
    \foreach \name / \y in {1,...,\numnodes}
        \node[hidden neuron] (H3-\name) at (\offsetb-\nodesep*\y, 6*\layersep) {};

    \foreach \name / \y in {1,...,\numnodes}
        \node[invisible neuron] (Z3B-\name) at (\offsetb-\nodesep*\y, 7*\layersep) {};

    \foreach \name / \y in {1,...,\numnodes}
        \node[invisible neuron] (Z4A-\name) at (\offsetb-\nodesep*\y, 10*\layersep) {};
        
    \foreach \name / \y in {1,...,\numnodes}
        \node[hidden neuron] (H4-\name) at (\offsetb-\nodesep*\y, 11*\layersep) {};
               
    \node[output neuron] at (\offsetb -\nodesep*\middlenode, 15 * \layersep )(O2) {};

    
  \node[hidden neuron, fill=blue!50] (H4fill-\ipos) at (\offsetb-\nodesep*\ipos, 11*\layersep) {};
  \node[hidden neuron, fill=darkgreen!50] (H3fill-\middlenode) at (\offsetb-\nodesep*\middlenode, 6*\layersep) {};


    \foreach \source in {1,...,\numnodes}
            \path[-] (I-\source) edge (H1-\source);
    \foreach \source[evaluate=\source as \dest using int(\source+1)] in {1,...,4}
            \path[-] (I-\source) edge (H1-\dest);
    \foreach \source[evaluate=\source as \dest using int(\source-1)] in {2,...,\numnodes}
            \path[-] (I-\source) edge (H1-\dest);

    \foreach \source in {1,...,\numnodes}
            \path[-] (H1-\source) edge (Z1-\source);
    \foreach \source[evaluate=\source as \dest using int(\source+1)] in {1,...,4}
            \path[-] (H1-\source) edge (Z1-\dest);
    \foreach \source[evaluate=\source as \dest using int(\source-1)] in {2,...,\numnodes}
            \path[-] (H1-\source) edge (Z1-\dest);

            \path[-] (Z1-\middlenode) -- node[auto=false]{\mydots} (Z3A-\middlenode);

    \foreach \source in {1,...,\numnodes}
            \path[-] (Z3A-\source) edge (H3-\source);
    \foreach \source[evaluate=\source as \dest using int(\source+1)] in {1,...,4}
            \path[-] (Z3A-\source) edge (H3-\dest);
    \foreach \source[evaluate=\source as \dest using int(\source-1)] in {2,...,\numnodes}
            \path[-] (Z3A-\source) edge (H3-\dest);
                        
    \foreach \source in {1,...,\numnodes}
            \path[-] (H3-\source) edge (Z3B-\source);
    \foreach \source[evaluate=\source as \dest using int(\source+1)] in {1,...,4}
            \path[-] (H3-\source) edge (Z3B-\dest);
    \foreach \source[evaluate=\source as \dest using int(\source-1)] in {2,...,\numnodes}
            \path[-] (H3-\source) edge (Z3B-\dest);
             
            \path[-] (Z3B-\middlenode) -- node[auto=false]{\mydots} (Z4A-\middlenode);            
    \foreach \source in {1,...,\numnodes}
            \path[-, draw=darkgreen!80, line width=0.4mm] (Z4A-\source) edge (H4-\source);
    \foreach \source[evaluate=\source as \dest using int(\source+1)] in {1,...,4}
            \path[-, draw=darkgreen!80, line width=0.4mm] (Z4A-\source) edge (H4-\dest);
    \foreach \source[evaluate=\source as \dest using int(\source-1)] in {2,...,\numnodes}
            \path[-, draw=darkgreen!80, line width=0.4mm] (Z4A-\source) edge (H4-\dest);

    \foreach \source in {1,...,\numnodes}
        \path[-, draw=darkgreen!80, line width=0.4mm] (H4-\source.north) edge (\offsetb -\nodesep*\middlenode, 13 * \layersep);
        
    \foreach \source in {1,...,\numnodes}
        \path[-] (\offsetb-\nodesep*0.5, 6*\layersep + 0.2 * \layersep* \middlenode - 0.2 * \layersep*\source) edge  (\offsetb + 0.25*\nodesep, 6 * \layersep);
        
        \path[-, draw=darkgreen!80, line width=0.4mm] (\offsetb-\nodesep*0.5, 6*\layersep) edge  (\offsetb + 0.2*\nodesep, 6 * \layersep);
        
        \path[-, draw=darkgreen!80, line width=0.4mm]   (\offsetb + 0.2*\nodesep, 6 * \layersep) edge (\offsetb +0.2* \nodesep, 11.1 * \layersep);
        \path[-{Latex[length=2mm,width=1.5mm]}, darkgreen!80, line width=0.4mm]  (\offsetb + 0.2*\nodesep, 11 * \layersep) edge  (O2.south);
                
    \foreach \source in {1,...,\numnodes}
        \path[-] (\offsetb-\nodesep*0.5, 1*\layersep + 0.2 * \layersep* \middlenode - 0.2 * \layersep*\source) edge (\offsetb +0.75*\nodesep, 1 * \layersep);
   
        \path[-]  (\offsetb +0.7*\nodesep, 1 * \layersep) edge (\offsetb +0.7*\nodesep, 11.3 * \layersep);
        \path[-{Latex[length=2mm,width=1.5mm]}]  (\offsetb +0.7*\nodesep, 11.2 * \layersep) edge (O2.south);
   
   	 \path[-{Latex[length=2mm,width=1.5mm]}, blue!80, line width=0.4mm]   (\offsetb -\nodesep*\middlenode, 12.9 * \layersep) edge (O2.south);
   
        \path[-] (\offsetb -\nodesep*\middlenode, 14 * \layersep) edge (O2);
   
        \path[-, draw=blue!80, line width=0.4mm] (H4-\ipos.north) edge (\offsetb -\nodesep*\middlenode, 13 * \layersep);
        \path[-, draw=darkgreen!80, line width=0.4mm] (H3-3) edge (Z3B-2);
        \path[-, draw=darkgreen!80, line width=0.4mm] (H3-3) edge (Z3B-3);
        \path[-, draw=darkgreen!80, line width=0.4mm] (H3-3) edge (Z3B-4);
        
   	 \node[annot,below of=I-5, node distance=0.5cm] {$x_0$};
   	 \node[annot,below of=I-4, node distance=0.5cm] {$x_1$};
   	 \node[annot,below of=I-3, node distance=0.5cm] {$x_2$};
   	 \node[annot,below of=I-2, node distance=0.5cm] {$x_3$};
   	 \node[annot,below of=I-1, node distance=0.5cm] {$x_4$};
            \node[annot,above of=O2, node distance=0.7cm] {Output};

   	 \node[annot, left of=I-5, node distance=1.3cm]  {Inputs $X$};
   	 \node[annot, left of=H1-5, node distance=1cm]  {$\phi_1(X)$};
	 \node[annot, left of=H3-5, node distance=1.8cm]  {$\phi_{\ell-1} \circ ... \circ \phi_1(X)$};
	 \node[annot, left of=H4-5, node distance=1.8cm]  {$\phi_{L-1} \circ ... \circ \phi_1(X)$};

   	 \node[annot] at (\offsetb -\nodesep*3.8, 13.7 * \layersep) {$L_{L-1}$};
   	 \node[annot] at (\offsetb -0.35*\nodesep, 8.5 * \layersep) {$L_{\ell-1}$};
   	 \node[annot] at (\offsetb + 0.25*\nodesep, 3.5 * \layersep) {$L_{1}$};
   
\end{tikzpicture}

\centering
\caption{\label{fig:schema_skip} A neural network with skip connections. The last layer $L$ can be partitioned as $\{L_1, L_2, \dots, L_{L-1} \}$, where $L_\ell$ processes the outputs of layer $\ell$. In green, the capacity propagation paths corresponding to $\kappa(L\to \ell, x_2)$. Compared to a network without skip connections, new paths appear due to $\kappa(L \to L_{\ell'}, x)$, $\ell' \geq \ell - 1$. The path that corresponds to $\kappa(L \to L_{\ell-1}, x_2)$ is shown on the right (green).}

\end{figure}

To facilitate the use of information from all layers -- which might represent different levels of abstraction -- a common practice is to use skip connections from intermediate layers to the output space. The last layer $L$ can then be partitioned as $\{L_1, L_2, \dots, L_{L-1} \}$ where $L_\ell$ corresponds to the part of the last layer that processes the outputs of layer $\ell$ (see Fig. \ref{fig:schema_skip}). The inputs of the last layer can be written as:

\begin{equation}
\underbrace{\phi(X)}_{\to L} = (\underbrace{\phi_1(X)}_{\to L_1}, \underbrace{\phi_{2} \circ \phi_1(X)}_{\to L_2}, ..., \underbrace{\phi_{L-1} \circ ... \circ \phi_1(X)}_{\to L_{L-1}}),
\end{equation}

\noindent where ``$\to$'' can be read as ``is processed by''. The last linear layer $L$ is therefore applied not only to the output of the last hidden layer, but to the outputs of every intermediate layer $1 \leq \ell \leq L-1$. Let us also partition the capacity allocated by the output layer to its inputs $\kappa(L \to L, x)$, according to  the above partition of the layer:

\begin{equation}
\kappa(L \to L, x) = (\underbrace{\kappa(L \to L_1, x)}_{\text{output of layer 1}}, \underbrace{\kappa(L \to L_2, x)}_{\text{output of layer 2}}, ..., \underbrace{\kappa(L \to L_{L-1}, x)}_{\text{output of layer $L-1$}}).
\end{equation}

\noindent Under the same scaling hypotheses as above, the capacity PDE becomes:

\begin{equation}\label{eq:capacity_PDE_skip}
\begin{cases}
\displaystyle
\frac{\partial\pi}{\partial t}(t, x) = \mathcal{D}\frac{\partial^2 \pi}{\partial x ^2}(t, x) + L \kappa(L \to L_{\ell(t)}, x),\\[6pt]
\pi(0, x) = 0,
\end{cases}
\end{equation}

\noindent The solution to this equation, and hence the propagated capacity $\kappa(L \to \ell, x)$, can again be derived explicitly:

\begin{equation}\label{eq:capacity_final_solution_multilayers}
\kappa(L \to \ell(t), x) := \pi(t, x) = L \int_{s=0}^t\int_{y=-\infty}^\infty  \frac{e^{-\frac{(x - y)^2}{4\mathcal{D}s}} }{\sqrt{4\pi \mathcal{D}s}} ~  \kappa(L \to L_{\ell(t - s)}, y) \mathrm{d}y \mathrm{d}s.
\end{equation}

\noindent For $\ell=1$ (i.e. $t=1$), the above equation describes the capacity allocation in the input space. Note that the capacity diverges with $L$ since the number of parameters in the last layer grows proportionally to the number of layer, so Eqs. (\ref{eq:capacity_PDE_skip})  and  (\ref{eq:capacity_final_solution_multilayers}) are ill-defined as such. One way around it is to consider the rescaled aggregated capacities $\kappa(L \to \ell, x) / L$, which remain finite as $L\to\infty$.

\subsection{Capacity associated with all the layers} 

So far we have only been able to provide a quantitative definition for the capacity allocated by very last linear layer in a network $\kappa(L \to \cdot~, x)$. Defining the capacities allocated by lower layers $\kappa(\ell \to \cdot~, x)$ does not appear to be straightforward in general -- but let us assume here that there is a way to achieve it, and denote $\kappa(\ell \to \ell~, x)$ the spatial capacity that layer $\ell\in[1, L]$ allocates to its own inputs. We then define the cumulated capacity allocated to some layer $\ell$, as the capacity allocated by \emph{all} the subsequent layers $\ell' \geq  \ell$ to $\ell$, which we note $\kappa(\{\ell, \dots, L\} \to \ell~, x)$. Correspondingly, we use the following compact notation in reverse layer time:

\begin{equation}
\pi^{\text{cum}}(t, x) := \kappa\left(\{\ell(t),\dots,L\} \to \ell(t), x\right).
\end{equation}

\noindent One can show that $\pi^{\text{cum}}$ follows an equation similar to Eq. (\ref{eq:capacity_PDE_skip}):

\begin{equation}\label{eq:capacity_PDE_all_layers}
\begin{cases}
\displaystyle
\frac{\partial\pi^{\text{cum}}}{\partial t}(t, x) = \mathcal{D}\frac{\partial^2 \pi^{\text{cum}}}{\partial x ^2}(t, x) + L \kappa(\ell(t) \to \ell(t), x),\\[6pt]
\pi^{\text{cum}}(0, x) = 0,
\end{cases}
\end{equation}

\noindent which again yields the explicit solution for $\kappa(\{\ell, \dots, L\} \to \ell~, x)$:

\begin{equation}\label{eq:capacity_final_solution_multilayers}
\kappa(\{\ell(t), \dots, L\} \to \ell(t)~, x) := \pi^{\text{cum}}(t, x) = L \int_{s=0}^t\int_{y=-\infty}^\infty  \frac{e^{-\frac{(x - y)^2}{4\mathcal{D}s}} }{\sqrt{4\pi \mathcal{D}s}} ~  \kappa(\ell(t-s) \to \ell(t-s), y) \mathrm{d}y \mathrm{d}s.
\end{equation}

\noindent The difference with Eq. (\ref{eq:capacity_final_solution_multilayers}) is that for $\ell = 1$ (i.e. $t=1$), the above expression now describes the capacity associated with every parameter in the network, projected in the input space -- i.e. the equivalent of what was obtained numerically in the linear case in \cite{donier2018capacity}.

\subsection{Providing inputs to intermediate layers}\label{sec:intermediate}

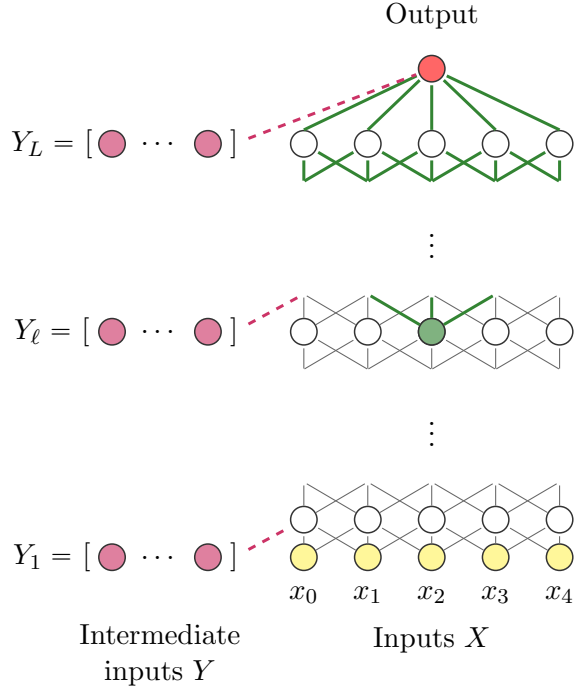
\begin{figure}[!t]
\def\layersep{0.5cm}
\def\numnodes{5}
\def\middlenode{3}
\def\nodesep{0.85}

\def\ipos{2}
\def \offsetb{0*\nodesep}

\definecolor{darkgreen}{RGB}{0, 102, 0}

\newcommand\mydots{\vdots}

\hspace*{4cm}
\begin{tikzpicture}[shorten >=1pt,->,draw=black!60, node distance=\layersep]
    \tikzstyle{every pin edge}=[<-,shorten <=1pt]
    \tikzstyle{neuron}=[circle,draw=black!80,minimum size=10pt,inner sep=0pt,line width=0.2mm]
    \tikzstyle{input neuron}=[neuron,fill=yellow!50];
    \tikzstyle{conditioning neuron}=[neuron,fill=purple!50];
    \tikzstyle{invisible neuron}=[neuron,minimum size=0pt];
    \tikzstyle{output neuron}=[neuron,fill=red!60];
    \tikzstyle{hidden neuron}=[neuron];
    \tikzstyle{annot} = [ text centered]

%
%

    \foreach \name / \y in {1,...,\numnodes}
        \node[input neuron] (I-\name) at (\offsetb-\nodesep*\y, 0) {};

        \node[conditioning neuron] (I-Y) at (\offsetb-1.5*\nodesep - \nodesep * \numnodes, 0) {};
        \node[conditioning neuron] (I-Y0) at (\offsetb-3*\nodesep - \nodesep * \numnodes, 0) {};

    \foreach \name / \y in {1,...,\numnodes}
        \node[hidden neuron] (H1-\name) at (\offsetb-\nodesep*\y, \layersep) {};
        
    \foreach \name / \y in {1,...,\numnodes}
        \node[invisible neuron] (Z1-\name) at (\offsetb-\nodesep*\y, 2*\layersep) {};       
          
    \foreach \name / \y in {1,...,\numnodes}
        \node[invisible neuron] (Z3A-\name) at (\offsetb-\nodesep*\y, 5*\layersep) {};
        
    \foreach \name / \y in {1,...,\numnodes}
        \node[hidden neuron] (H3-\name) at (\offsetb-\nodesep*\y, 6*\layersep) {};
        
        \node[conditioning neuron] (H3-Y) at (\offsetb-1.5*\nodesep - \nodesep * \numnodes, 6*\layersep) {};
        \node[conditioning neuron] (H3-Y0) at (\offsetb-3*\nodesep - \nodesep * \numnodes, 6*\layersep) {};

    \foreach \name / \y in {1,...,\numnodes}
        \node[invisible neuron] (Z3B-\name) at (\offsetb-\nodesep*\y, 7*\layersep) {};

    \foreach \name / \y in {1,...,\numnodes}
        \node[invisible neuron] (Z4A-\name) at (\offsetb-\nodesep*\y, 10*\layersep) {};
        
    \foreach \name / \y in {1,...,\numnodes}
        \node[hidden neuron] (H4-\name) at (\offsetb-\nodesep*\y, 11*\layersep) {};
        
        \node[conditioning neuron] (H4-Y) at (\offsetb-1.5*\nodesep - \nodesep * \numnodes, 11*\layersep) {};
        \node[conditioning neuron] (H4-Y0) at (\offsetb-3*\nodesep - \nodesep * \numnodes, 11*\layersep) {};
               
    \node[output neuron, above of=H4-\middlenode, node distance=1cm] (O2) {};

    
  \node[hidden neuron, fill=darkgreen!50] (H3fill-\middlenode) at (\offsetb-\nodesep*\middlenode, 6*\layersep) {};


    \foreach \source in {1,...,\numnodes}
            \path[-] (I-\source) edge (H1-\source);
    \foreach \source[evaluate=\source as \dest using int(\source+1)] in {1,...,4}
            \path[-] (I-\source) edge (H1-\dest);
    \foreach \source[evaluate=\source as \dest using int(\source-1)] in {2,...,\numnodes}
            \path[-] (I-\source) edge (H1-\dest);

    \foreach \source in {1,...,\numnodes}
            \path[-] (H1-\source) edge (Z1-\source);
    \foreach \source[evaluate=\source as \dest using int(\source+1)] in {1,...,4}
            \path[-] (H1-\source) edge (Z1-\dest);
    \foreach \source[evaluate=\source as \dest using int(\source-1)] in {2,...,\numnodes}
            \path[-] (H1-\source) edge (Z1-\dest);

            \path[-] (Z1-\middlenode) -- node[auto=false]{\mydots} (Z3A-\middlenode);

    \foreach \source in {1,...,\numnodes}
            \path[-] (Z3A-\source) edge (H3-\source);
    \foreach \source[evaluate=\source as \dest using int(\source+1)] in {1,...,4}
            \path[-] (Z3A-\source) edge (H3-\dest);
    \foreach \source[evaluate=\source as \dest using int(\source-1)] in {2,...,\numnodes}
            \path[-] (Z3A-\source) edge (H3-\dest);
                        
    \foreach \source in {1,...,\numnodes}
            \path[-] (H3-\source) edge (Z3B-\source);
    \foreach \source[evaluate=\source as \dest using int(\source+1)] in {1,...,4}
            \path[-] (H3-\source) edge (Z3B-\dest);
    \foreach \source[evaluate=\source as \dest using int(\source-1)] in {2,...,\numnodes}
            \path[-] (H3-\source) edge (Z3B-\dest);
             
            \path[-] (Z3B-\middlenode) -- node[auto=false]{\mydots} (Z4A-\middlenode);            
    \foreach \source in {1,...,\numnodes}
            \path[-, draw=darkgreen!80, line width=0.4mm] (Z4A-\source) edge (H4-\source);
    \foreach \source[evaluate=\source as \dest using int(\source+1)] in {1,...,4}
            \path[-, draw=darkgreen!80, line width=0.4mm] (Z4A-\source) edge (H4-\dest);
    \foreach \source[evaluate=\source as \dest using int(\source-1)] in {2,...,\numnodes}
            \path[-, draw=darkgreen!80, line width=0.4mm] (Z4A-\source) edge (H4-\dest);

    \foreach \source in {1,...,\numnodes}
        \path[-, draw=darkgreen!80, line width=0.4mm] (H4-\source.north) edge (O2);
                 
        \path[-, draw=darkgreen!80, line width=0.4mm] (H3-3) edge (Z3B-2);
        \path[-, draw=darkgreen!80, line width=0.4mm] (H3-3) edge (Z3B-3);
        \path[-, draw=darkgreen!80, line width=0.4mm] (H3-3) edge (Z3B-4);
        
   	 \node[annot,below of=I-5, node distance=0.5cm] (x0) {$x_0$};
   	 \node[annot,below of=I-4, node distance=0.5cm] {$x_1$};
   	 \node[annot,below of=I-3, node distance=0.5cm] (x2) {$x_2$};
   	 \node[annot,below of=I-2, node distance=0.5cm] {$x_3$};
   	 \node[annot,below of=I-1, node distance=0.5cm] {$x_4$};
            \node[annot,above of=O2, node distance=0.7cm] {Output};

   	 \node[annot, below of=x2, node distance=0.6cm]  (inp) {Inputs $X$};
	 \node[annot] at (\offsetb-2.25*\nodesep - \nodesep * \numnodes, -2.6*\layersep)  {\begin{tabular}{c}Intermediate \\inputs $Y$\end{tabular}};
   	 \node[annot, left of=I-Y0, node distance=0.8cm]  {$Y_1 =$ [};
	 \node[annot, left of=H3-Y0, node distance=0.8cm]  {$Y_{\ell} =$ [};
	 \node[annot, left of=H4-Y0, node distance=0.8cm]  {$Y_{L} =$ [};
	 
   	 \node[annot, right of=I-Y, node distance=0.35cm]  (rbraceI) {]};
   	 \node[annot, right of=H3-Y, node distance=0.35cm]  (rbrace3) {]};
   	 \node[annot, right of=H4-Y, node distance=0.35cm]  (rbrace4) {]};

      \path[-, dashed, draw=purple!80, line width=0.4mm]  (rbraceI) edge (H1-5);
      \path[-, dashed, draw=purple!80, line width=0.4mm]  (rbrace3) edge (Z3B-5);
      \path[-, dashed, draw=purple!80, line width=0.4mm]  (rbrace4) edge (O2);

	 \path[-] (I-Y0) -- node[auto=false]{\dots} (I-Y);
	 \path[-] (H3-Y0) -- node[auto=false]{\dots} (H3-Y);
	 \path[-] (H4-Y0) -- node[auto=false]{\dots} (H4-Y);

\end{tikzpicture}

\centering
\caption{\label{fig:schema_conditioning} A neural network with both regular inputs ($X$, in yellow) and intermediate inputs provided at every layers ($\{Y_1, \dots, Y_L\}$, in purple).  In green, the capacity propagation paths corresponding to $\kappa(L\to \ell, x_2)$. The ``capacity leak'' which goes to the space of $Y$ is represented by the dashed purple lines. Any capacity that is allocated to $\{Y_1, \dots, Y_L\}$ is therefore not allocated to $X$. }

\end{figure}

Conversely to skip connections, it is common practice to provide some additional inputs to intermediate layers, e.g. by concatenating them or summing them with the layer inputs (see Figure \ref{fig:schema_conditioning}). This has been done in particular in the case of conditioned models, where some base input $X$ is provided to the first layer only, but some conditioning $Y_\ell$ is provided to every layer throughout the network \cite{van2016wavenet}, or more recently in a GAN framework \cite{karras2018style}. In this case, the transformation operated by a single layer can be written as:

\begin{equation}\label{eq:layer_leak}
  X_\ell = \phi_\ell(X_{\ell-1}, Y_\ell) := f_\ell(W_\ell^TX_{\ell-1} + \gamma_\ell Y_\ell),
\end{equation}

\noindent where $Y_\ell$ is the conditioning at layer $\ell$.\footnote{For simplifying the notations, we add $\gamma _\ell Y_\ell$ to $W_\ell^TX_{\ell-1}$ directly instead of some linear transformation $V_\ell^TY_\ell$, but this case can be treated similarly.} We consider the following scaling relations:

\begin{equation}
\begin{cases}
\begin{aligned}
& \text{Residual capacity:} & \epsilon ~\propto ~1 / L \\
&\text{Residual weights:}& \epsilon^W := \sqrt{\epsilon} ~\propto~ 1 / \sqrt{L}\\
&\text{Intermediate inputs:}&\gamma ~\propto ~\sqrt{\epsilon}~\propto~ 1 / \sqrt{L}\\
\end{aligned}
\end{cases}
\end{equation}

\noindent According to the last point, we note $\gamma_\ell := \sqrt{\alpha_\ell\epsilon}$ with $\alpha_\ell>0$ of order unity. The discrete propagation equation from layer $\ell$ to layer $\ell-1$ then reads:

\begin{equation}
\begin{bmatrix}
\kappa^{\ell-1}\\
\kappa^{Y_\ell}
  \end{bmatrix} = D_\ell.\kappa^\ell,\quad \text{with}\quad
D_\ell
  =
\begin{bmatrix}
    (1 - \alpha_\ell\epsilon) W_\ell\circ W_\ell \\
    \alpha_\ell\epsilon \mathbb{I}
  \end{bmatrix} + O\left(\epsilon^2\right).
\end{equation}

\noindent The interpretation of this equation is that not all the capacity $\kappa^\ell$ is propagated down to the previous layer, as a small fraction ($\kappa^{Y_\ell}~\propto ~ \alpha_\ell\epsilon$) is being allocated to $Y_\ell$ at each layer. In the continuous limit, this turns into the following PDE:

\begin{equation}\label{eq:layer_leak_PDE}
\begin{cases}
\displaystyle
\frac{\partial\pi}{\partial t}(t, x) = \mathcal{D}\frac{\partial^2 \pi}{\partial x ^2}(t, x) - \alpha(t) \pi(t, x),\\[6pt]
\pi(0, x) = \kappa^L(x),
\end{cases}\quad \text{with}\quad 
\begin{cases}
 \kappa^X(x) = \pi(1, x)\\
\kappa^Y(x) =  \int_0^1 \alpha(t) \pi(t, x) \mathrm{d}t
\end{cases}
\end{equation}

\noindent where we have noted $\kappa^X$ the capacity allocated in the space of $X$ and $\kappa^Y$ the sum of the capacities allocated in the space of $Y$. This is akin to adding a ``capacity leak'' that depletes $\pi(t)$ and which goes straight to the $Y$ space, with $\sum \kappa^X + \sum\kappa^Y = \sum\kappa^L$ (conservation of total capacity through the network). 

Note that for the above limit to be non-trivial and $\kappa^X$ to not be vanishingly small, the scale of the intermediate inputs needs to be of order $1 / \sqrt{L}$, which is not a condition that is typically enforced in the literature -- an experiment worth trying.

\subsection{The bias term}

So far we have omitted the bias in the layers and used linear rather than affine transformations. What happens if we add it back? This is similar to Eq. (\ref{eq:layer_leak}), where the intermediate inputs $Y_l$ are replaced by the parameter vectors $B_l$:

\begin{equation}\label{eq:layer_bias}
 X_\ell = \phi_\ell(X_{\ell-1}) = f_\ell(W_\ell^TX_{\ell-1} + \beta_\ell B_\ell),
\end{equation}

\noindent where $\beta_\ell$ is the scale applied to the bias term. Similarly to the previous section, we use the following scaling relations:

\begin{equation}
\begin{cases}
\begin{aligned}
& \text{Residual capacity:} & \epsilon ~\propto ~1 / L \\
&\text{Residual weights:}& \epsilon^W := \sqrt{\epsilon} ~\propto~ 1 / \sqrt{L}\\
&\text{Bias:}&\beta ~\propto ~\sqrt{\epsilon}~\propto~ 1 / \sqrt{L}
\end{aligned}
\end{cases}
\end{equation}

\noindent and write $\beta_\ell = \sqrt{\alpha_\ell\epsilon}$ accordingly with $\alpha_\ell$ of order unity. For the sake of simplicity, let us consider the case where the bias is constant for each layer $B_\ell = \mathbbm{1}$, such that we recover exactly the equations from the previous section with $Y = \mathbbm{1}$. The PDE for $\pi(t)$ is therefore the same as Eq. (\ref{eq:layer_leak_PDE}):

\begin{equation}\label{eq:capacity_PDE_bias}
\begin{cases}
\displaystyle
\frac{\partial\pi}{\partial t}(t, x) = \mathcal{D}\frac{\partial^2 \pi}{\partial x ^2}(t, x) - \alpha(t) \pi(t, x),\\[6pt]
\pi(0, x) = \kappa^L(x),
\end{cases}\quad \text{with}\quad 
 \kappa^X(x) = \pi(1, x),
 \end{equation}

\noindent with the only difference that the ``leaked'' capacity $\int_0^1 \alpha(t) \pi(t, x) \mathrm{d}t$ now goes to constant components rather than some conditioning $Y$. This suggests that similarly to the conditioning, the biases should be scaled in an appropriate way for the PDE to make sense, and the capacity at the input layer not to vanish -- more precisely, the bias terms should be scaled down as $1 / \sqrt{L}$ as the number of layers $L$ grows large, similarly to the residual weights.

\subsection{Dilated convolutions} 

The case of dilated convolutions can be analyzed, under the same scaling hypotheses as in Section \ref{sec:vanilla}, by replacing $\mathcal{D} \to \mathcal{D}(t)$ in the capacity PDE (Equation (\ref{eq:capacity_PDE})) and $\mathcal{D}t \to V(t) := \int_{s=0}^t \mathcal{D}(s) \mathrm{d}s$ in the integral form (Equation (\ref{eq:capacity_final_solution})). Indeed, increasing the dilation rate results in a (quadratically) higher diffusion rate between two layers.  For example, in the case of layers with exponentially increasing receptive fields $\mathcal{D}(t) = e^{\alpha (1 - t)}$, this gives:

\begin{equation}
V(1) = \frac{1}{\alpha} \left(e^\alpha - 1\right). 
\end{equation}

\noindent If the dilation ratio per layer is $\lambda$ such that the receptive field of the last layer is $R_L ~\propto~ \lambda^L$, then the ratio between the uppermost and lowermost diffusivities is $e^{\alpha} = \mathcal{D}(1) / \mathcal{D}(0) = \lambda^{2(L - 1) }$, which gives

\begin{equation}
V(1) = \frac{\lambda^{2L- 2}}{(2L - 2) \mathrm{log}(\lambda)}
\end{equation}

\noindent The spatial capacity in the input space corresponding to a one-hot output capacity $\kappa^L(x) = \delta(x - x_0)$ then reads:

\begin{equation}\label{eq:capacity_dirac_dilated}
\kappa^1(x) =\frac{e^{-\frac{(x-x_0)^2}{4V(1)}}}{\sqrt{4\pi V(1)}}.
\end{equation}

\noindent The capacity is therefore non-vanishing for:

\begin{equation}
x - x_0 = O\left(\sqrt{V(1)}\right) = O\left( \frac{\lambda^L}{\sqrt{(2L - 2)\lambda \mathrm{log}(\lambda)}}\right) = O\left(\frac{R_L}{\sqrt{L}}\right).
\end{equation}

\noindent In that case, the effective receptive field therefore almost scales with the largest receptive field $R_L$, up to some $1 / \sqrt{L}$ scaling factor.\footnote{Note that this factor is a consequence of the choice that was made to have a smooth exponential growth for the dilation. Other choices -- for example, a growth with fewer but larger steps -- would lead to different correcting factors.}

\subsection{Multi-channel networks}

So far, we have only studied single channel networks for the sake of simplicity, as they give rise to simple PDEs on 1-dimensional inputs. The case of multi-channel networks can however be treated in a similar way, by considering $C$ parallel processes that are entangled when passing through the layers. More precisely, the discrete propagation equation from layer $\ell$ to layer $\ell-1$ reads:

\begin{equation}
\begin{bmatrix}
\kappa^{\ell-1}_1\\
\vdots\\
\kappa^{\ell - 1}_{C}
  \end{bmatrix} = D_\ell.
  \begin{bmatrix}
\kappa^{\ell}_1\\
\vdots\\
\kappa^{\ell}_{C}
  \end{bmatrix}, \quad \text{where} \quad 
  D_\ell :=  \mathbb{I} + \epsilon \Delta_\ell \quad \text{and} \quad
  \Delta_\ell = 
  \left[
  \begin{matrix}
  \Delta_{\ell, 11} & \hdots &   \Delta_{\ell, 1C} \\
  \vdots & & \vdots\\
  \Delta_{\ell, C1} & \hdots &   \Delta_{\ell, CC} 
  \end{matrix}
  \right],
\end{equation}

\noindent where $1\dots C$ are the channel indices, $\kappa^\ell_c$ the capacity allocated to the $c$-th channel of layer $\ell$ and $\Delta_{\ell, cc'}$ represents the residual capacity propagation from  channel $c'$ of layer $\ell$ to channel $c$ of layer $\ell-1$. 

The above equation can be collapsed along the channel dimension: if we note $\kappa^\ell = \sum_{c=1}^{C} \kappa^{\ell}_c$ the capacity allocated to some spatial position across all channels, then the propagation equation for $\kappa^\ell$ can be written under the form:

\begin{equation}
\begin{bmatrix}
\kappa^{\ell-1}_1\\
\vdots\\
\kappa^{\ell - 1}_{C}
  \end{bmatrix} = 
    \begin{bmatrix}
\kappa^{\ell}_1\\
\vdots\\
\kappa^{\ell}_{C}
  \end{bmatrix}
+ \epsilon
    \left[
  \begin{matrix}
  \hat{\Delta}_{\ell, 1} (\kappa^{\ell})\\
    \vdots \\
    \hat{\Delta}_{\ell, C} (\kappa^{\ell})
    \end{matrix}
      \right]
  . \kappa^{\ell},
\end{equation}

\noindent where $\hat{\Delta}_{\ell, c} (\kappa^{\ell})$ is some $\kappa_\ell$-dependent average of $\{   \Delta_{\ell, cc'} \}_{c'=1\dots C}$ (which has therefore the same order of magnitude as individual $\Delta_{\ell, cc'}$'s). The evolution equation for $\kappa^\ell$ can therefore be obtained by summing over the channels:

\begin{equation}
\kappa^{\ell-1} = \left(\mathbb{I} + \epsilon
\sum_{c=1}^C   \hat{\Delta}_{\ell, c} (\kappa^{\ell})\right)
  . \kappa^{\ell}.
\end{equation}

\noindent This is reminiscent of the discrete propagation equation in the case of single channel models (cf. \cite{donier2018nonlinear}), for which the appropriate scaling was to have $\Delta$ of order $1$, which corresponds here to having $\sum_{c=1}^C   \hat{\Delta}_{l, c} (\kappa^{\ell})$ of order 1, and therefore (assuming that the scale of the weights are uniform across channels) having $\hat{\Delta}_{\ell, c} (\kappa^{\ell})$ of order $1 / C$ -- which, from its definition, means having $\Delta_{\ell, cc'}$ of order $1 / C$. In summary, multi-channels networks require the following scaling relations:

\begin{equation}
\begin{cases}
\begin{aligned}
& \text{Residual capacity:} & \epsilon ~\propto ~1 / (C L) \\
&\text{Residual weights:}& \epsilon^P := \sqrt{\epsilon} ~\propto~ 1 / \sqrt{C L}\\
\end{aligned}
\end{cases}
\end{equation}

\noindent This is particularly interesting as we recover the well-known scaling of the weights as the inverse square root of the number of channels introduced in \cite{glorot2010understanding}, that is nowadays commonly used to fix the scale of the weights in the initialization phase -- and commonly known as the ``Xavier'' initializer.\footnote{To be precise, \cite{glorot2010understanding} propose to scale the weights as $\sim 1 / \sqrt{C_{\ell-1} + C_\ell}$ where $C_\ell$ is the number of channels at layer $l$. Here we use the same number of channels $C$ throughout the network, in which case the Xavier scaling is equivalent to scaling the weights as $\sim 1 / \sqrt{C}$.}





\subsection{Multi-dimensional inputs}

While our study has focussed on 1-dimensional data so far, the same can be applied to inputs of dimension $d\geq 1$ (like images). Indeed, multi-dimensional tensors can always be flattened as a 1-dimensional vector. In particular, the $1 / \sqrt{L}$ scaling for the weights still holds. The main difference is the sparsity of the weights matrix $W$, which now has $O(r^d)$ non-zero values for convolutional filters of size $r$. Therefore, the scale of the off-diagonal weights should be scaled down by a factor $\sqrt{r^{d-1}}$ compared to the 1-D convolutions, so that the corresponding capacity weights are scaled down by a factor $r^{d-1}$, thus ensuring that the residual part has the same aggregate weight. The multi-dimensional PDE reads:

\begin{equation}\label{eq:capacity_PDE_multidim}
\begin{cases}
\displaystyle
\frac{\partial\pi}{\partial t}(t, x) = \sum_{i, j = 1}^d\mathcal{D}_{ij}\frac{\partial^2 \pi}{\partial x_i\partial x_j}(x, t),\\[6pt]
\pi(0, x) = \kappa^L(x),
\end{cases}
\end{equation}

\noindent and the scaling relations become, according to the above:

\begin{equation}
\begin{cases}
\begin{aligned}
& \text{Residual capacity:} & \epsilon ~\propto ~1 / (r^d L) \\
&\text{Residual weights:}& \epsilon^W := \sqrt{\epsilon} ~\propto~ 1 / \sqrt{r^d L}\\
\end{aligned}
\end{cases}
\end{equation}

\subsection{Recurrent networks}

The case of recurrent networks with residual state transformations is similar to networks with inputs provided to intermediate layers studied in Section \ref{sec:intermediate}, but where the depth $\ell$ is replaced by the time $t$. Accordingly, the side inputs $Y$ are replaced by the sequential inputs $Y_t$ and the hidden outputs $X_\ell$ are replaced by the recurrent states $X_t$ -- and the network input $X$ is replaced by the initial state $X_0$ (to get a mental picture, imagine applying a $90^\degree$ clockwise rotation to the network from Figure \ref{fig:schema_conditioning}, to turn it into a shallow recurrent network). Therefore, the same propagation equation arises (cf Eq. \ref{eq:layer_leak_PDE}), but this time with $t$ representing the actual time:

\begin{equation}
\begin{cases}
\displaystyle
\frac{\partial\pi}{\partial t}(t, x) = \mathcal{D}\frac{\partial^2 \pi}{\partial x ^2}(t, x) - \alpha(t) \pi(t, x),\\[6pt]
\pi(0, x) = \kappa^L(x),
\end{cases}\quad \text{with}\quad 
\begin{cases}
 \kappa^{X_0}(x) = \pi(1, x)\\
\mathrm{d}\kappa^{Y_t}(x) =  \alpha(t) \pi(t, x) \mathrm{d}t
\end{cases}
\end{equation}

\noindent where $\mathrm{d}\kappa^{Y_t}$ is the capacity allocated to the input at time $t$. Note that it is now infinitesimal, as a finite capacity now is being allocated to an infinite number of inputs. The scaling relations are similar to that of residual networks, if we note $N$ the number of inputs (the analog of $L$ for deep residual networks):

\begin{equation}
\begin{cases}
\begin{aligned}
& \text{Residual capacity:} & \epsilon ~\propto ~1 / N \\
&\text{Residual weights:}& \epsilon^W := \sqrt{\epsilon} ~\propto~ 1 / \sqrt{N}.\\
\end{aligned}
\end{cases}
\end{equation}

\noindent Note that we have implicitly assumed that the state transformations were done using local weights (convolutional-type) in order to obtain a local propagation, which might not be realistic (or at least, which has never been tried to our knowledge). Still, this configuration ought to be tried as achieving long memory in recurrent networks has been a long time challenge.

\section{Conclusion}

We have formulated what we called the \emph{neural network scaling conjecture}, which hypothesizes that deep neural networks can only avoid the shattering problem if the equation that describes the capacity propagation through their layers has a non-degenerate continuous limit when the number of layers tends to infinity. If this is true, this imposes a number of scaling relations between the depth of a network and the scale of its various parameters for it to function properly. Several common types of networks have been studied, including architectures with skip connections, conditioning along the network, dilations, recurrence, etc. Interestingly, we have found that a similar requirement emerges in all the cases: for a networks with $L\gg 1$ layers, any weights other than those that simply pass the inputs through must scale as $1/\sqrt{L}$ -- so that the squares of these weights, which define the diffusion of capacity, scale as $1 / L$. For recurrent networks with $N$ sequential inputs, the requirements are similar and the weights must scale as $1 / \sqrt{N}$. In the multi-channels case, we have recovered the condition required by the Xavier initialization, and we have briefly considered the extension to multi-dimensional data. This mix of known and new results is encouraging, and shows that the capacity approach may provide a simple and unified framework to think about the problem of scaling up deep neural networks. Of course, our results still need to be generalized to more generic types of activation functions -- and be put under the scrutiny of more empirical tests.


\section*{Acknowledgements} The author would like to thank Pierre Baqu\'{e}, Marc Sarfati and Antoine Tilloy for their very useful comments on the manuscript.

\bibliography{Scaling_up_deep_networks}{}
\bibliographystyle{plain}

\end{document}